\documentclass[9pt,conference]{IEEEtran}
\usepackage[pdftex]{graphicx}
\usepackage{amsmath}
\usepackage{graphicx, subfigure}

\begin{document}

\title{SIFT Vs SURF: Quantifying the Variation in Transformations}

\author{\IEEEauthorblockN{Siddharth Srivastava}
\IEEEauthorblockA{Department of Electrical Engineering, Indian Institute of Technology, Delhi\\ \\
eez127506@iitd.ac.in}}
\maketitle

\begin{abstract}
This paper studies the robustness of SIFT and SURF against different image transforms (rigid body, similarity, affine and projective) by quantitatively analyzing the variations in the extent of transformations. Previous studies have been comparing the two techniques on absolute transformations rather than the specific amount of deformation caused by the transformation. The paper establishes an exhaustive empirical analysis of such deformations and matching capability of SIFT and SURF with variations in matching parameters and the amount of tolerance. This is helpful in choosing the specific use case for applying these techniques. \\

\end{abstract}


\section{Introduction}
Natural images may suffer from many deformations like rotation, scale, shear, viewpoint etc. Geometric transformations are used to explain these deformations on images. There are primarily four categories transformations, namely Rigid Body transformation, Similarity Transformation, Affine Transformation and Perspective Transformation, with perspective transformation being the most general of the four.  Since these transformations are usually very common in real world images, it becomes important to be able analyze the images while minimizing the deformations introduced while capturing them. This process is achieved by describing an image as a set of features which uniquely identifies the image or acts as fingerprint for the image. Many techniques have been proposed and compared with each other for this purpose \cite{khan2011sift, juan2009comparison, mikolajczyk2005performance}. SIFT \cite{lowe1999object} and SURF \cite{bay2006surf} are two such widely used feature detection techniques. 

The primary aim of this paper has been to perform various transformations on datasets of images and study the matching capability of SIFT and SURF features. The paper is organized as follows. Section \ref{sec:methodology} describes the methodology used in this paper with regard to the transformations considered, the feature extraction techniques and the matching algorithm. This is followed by a discussion of the results obtained in Section \ref{sec:results} followed by the conclusion in Section \ref{sec:conclusion}.

\section{Methodology\label{sec:methodology}}
This section discusses the methodology adopted in our work. We begin by presenting a discussion of the dataset used and the motivation for choosing the dataset. Subsequently, we detail the transformations considered as applied to the dataset. We then present a brief discussion of the feature extraction techniques and the matching algorithm used. 

\begin{figure}[h]
\centering
\subfigure[]
{
\includegraphics[width=0.25\columnwidth]{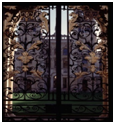}
\label{fig:1a}
}
\subfigure[]
{
\includegraphics[width=0.25\columnwidth]{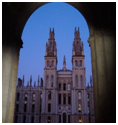}
\label{fig:1b}
}
\subfigure[]
{
\includegraphics[width=0.25\columnwidth]{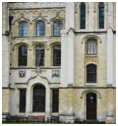}
\label{fig:1c}
}
\subfigure[]
{
\includegraphics[width=0.25\columnwidth]{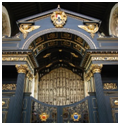}
\label{fig:1d}
}
\subfigure[]
{
\includegraphics[width=0.25\columnwidth]{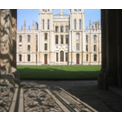}
\label{fig:1e}
}
\caption{The dataset used for the study (derived from Oxford Buildings Dataset)}
\label{fig:OxfordDataset}
\end{figure}

A reference dataset consisting of 10 images from the Oxford buildings dataset \cite{philbin2007oxford} has been chosen. The size of the images in the original dataset is either 1024x768 or 768x1024. For computational efficiency the images have been scaled down by 50\% along both the dimensions.
The images have been chosen to test the SIFT and SURF with differing category of content in the images. While the dataset consists of the buildings, each image has been chosen keeping certain parameters in mind. Fig. \ref{fig:1a} has a lot of fine details, Fig \ref{fig:1b} and Fig \ref{fig:1c} are of the same building under different lighting and viewing conditions. Fig \ref{fig:1c} is the front view of a normal building and Fig \ref{fig:1d} has textural details.  The reason for choosing such images has been to incorporate the above mentioned factors for testing the robustness of the feature matching techniques against the transformations as discussed in previous section.

\subsection{Image Transformations}
Following transformations were applied to the images to generate a cumulative dataset for testing. 

\begin{figure}[h]
\centering
\subfigure[]
{
\includegraphics[width=0.25\columnwidth]{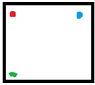}
\label{fig:2a}
}
\subfigure[]
{
\includegraphics[width=0.25\columnwidth]{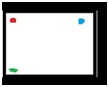}
\label{fig:2b}
}
\subfigure[]
{
\includegraphics[width=0.25\columnwidth]{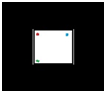}
\label{fig:2c}
}
\subfigure[]
{
\includegraphics[width=0.25\columnwidth]{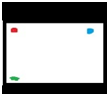}
\label{fig:2d}
}
\subfigure[]
{
\includegraphics[width=0.25\columnwidth]{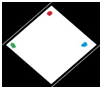}
\label{fig:2e}
}
\subfigure[]
{
\includegraphics[width=0.25\columnwidth]{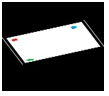}
\label{fig:2f}
}
\caption{Affine Transformations applied to the images.}
\label{fig:AffineTransforms}
\end{figure}

\begin{figure}[h]
\centering
\subfigure[]
{
\includegraphics[width=0.25\columnwidth]{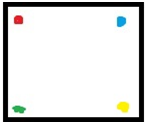}
\label{fig:3a}
}
\subfigure[]
{
\includegraphics[width=0.25\columnwidth]{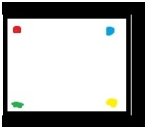}
\label{fig:3b}
}
\subfigure[]
{
\includegraphics[width=0.25\columnwidth]{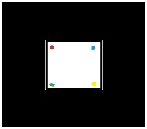}
\label{fig:3c}
}
\subfigure[]
{
\includegraphics[width=0.25\columnwidth]{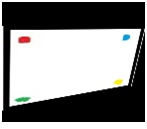}
\label{fig:3d}
}
\subfigure[]
{
\includegraphics[width=0.25\columnwidth]{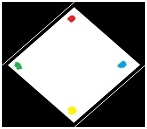}
\label{fig:3e}
}
\subfigure[]
{
\includegraphics[width=0.25\columnwidth]{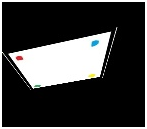}
\label{fig:3f}
}
\caption{Perspective Transformation applied to the images.}
\label{fig:PerspectiveTransforms}
\end{figure}

\begin{enumerate}

\item	Scaling:   The original images were scaled in the following ratios with respect to the reference images: 0.125, 0.25, 0.5, 0.75, 2 and 4. 
\item	Rotation: The original image has been rotated with the following angles in anticlockwise direction (degrees): 10, 20, 30, 40, 50, 90, and 180. 
\item	Similarity Transform: On the rotation images generated previously, scaling was applied in the following ratio: 0.25, 0.5, 2 and 4. This resulted in 28 similarity transform images 
\item	Affine Transform: Each reference image was transformed with 5 different affine transformations. First, an affine transform was obtained transforming the top left, top right and bottom left corners of the reference image to different locations in the target image.  The obtained affine transform matrix was then applied to the entire image to obtain the affine transformed image. 

The transformations applied are shown in Fig \ref{fig:AffineTransforms}. Fig \ref{fig:2a} is the reference image with red, blue and green patches indicating the corners considered for getting the transform matrix.  The corresponding corners are also shown in the affine transformed images in Fig \ref{fig:2b} - Fig \ref{fig:2f}. As can be seen from the transformed images, the parallel lines from the reference image are preserved in all the transformed images.

\item Perspective Transform: Each reference image was transformed with 5 different perspective transformation matrices. Though affine transform is a special case of perspective transform, to study the effect of both affine and perspective transformations quantitatively ,the three points considered in affine transformation were kept the same in the perspective transformation but the fourth point (bottom right corner of the image) was varied in these transformations.

Fig \ref{fig:3a} is the reference image indicating the corner points with colored patches. Comparing it with Fig \ref{fig:AffineTransforms}, we can see that Fig \ref{fig:3b}, \ref{fig:3c} and \ref{fig:3e} correspond to Fig \ref{fig:2b}, \ref{fig:2c} and \ref{fig:2e} respectively. In these cases, perspective transform was obtained by keeping the transformed location of the fourth corner point aligned proportionally with the three corners of the affine transformation.  This shows that affine transform is indeed a specific case of perspective transform.

Another point observed is that the perspective transform only preserves the straight lines.  As can be seen from Fig \ref{fig:3d} and Fig \ref{fig:3f}, the parallelism among the lines has been lost. 
\end{enumerate}

\begin{figure}[h]
\centering
\includegraphics[width=\columnwidth]{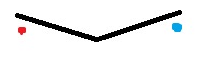}
\caption{Bending the line.}
\label{fig:4}
\end{figure}

\begin{figure}[h]
\centering
\includegraphics[width=\columnwidth]{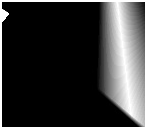}
\caption{Bending the line.}
\label{fig:5}
\end{figure}

An attempt was to made to deform straight lines by applying a transform which bends the line joining the top left and top right corner along the center of the line as shown in Fig \ref{fig:4}. The transformed image is shown in the Fig \ref{fig:5}.

As can been seen from Fig \ref{fig:5}, the transformed image has visual loss in terms of intensity changes as well as the deformation is not at all close to the expected deformation of Fig \ref{fig:4}.  The attempt to recover Fig \ref{fig:3a} by applying the inverse perspective transform on Fig \ref{fig:5} failed. 

\subsection{Feature Extraction\label{subsec:featureextraction}}
This section discusses the feature extraction algorithms used.

\subsubsection{Scale Invariant Feature Transform (SIFT)}

The SIFT algorithm is described in brief as follows:
\begin{enumerate}
\item	SIFT applies Gaussian filter to the image at various scales which are called octaves. Each octave is a collection of successively blurred images. Octaves differ with each other in the scale (usually 1/2 of previous octave). This is called scale space analysis. In second step, it calculates Difference of Gaussian (DoG) from successively blurred image which provides it scale invariance.  
\item For finding the keypoints, it finds maxima and minima in DoG images and then finds sub-pixel minima and maxima from them using Taylor's series expansion.
\item	Next, the erroneous key-points are eliminated by thresholding. So, it aims at finding the corner points for stronger keypoints.
\item	Then the orientation is assigned to each keypoint within a region depending upon the scale of the image. Since the orientation of each sub region is adjusted against the orientation of the keypoint's region (by subtraction), rotation invariance is achieved.
\item	Feature estimation: By considering a 16x16 region around it and the orientation is calculated for each 4x4 region in it. A histogram is plotted with 8 bins but the assigned bin for each orientation is dependent upon the distance of the region from the key-point. This is achieved with the help of a Gaussian weighted function which also provides robustness to deformations and translation. Since there are 4x4 regions and 8 bins, SIFT calculates 128 dimensional feature vector.
\end{enumerate}

\subsubsection{Speeded Up Robust Features (SURF)}

SURF is also a feature extraction technique which claims to be more robust and faster than SIFT. The algorithm highlighting the key difference from the SIFT as described above are described below:

\begin{enumerate}
\item	SURF uses Integral images for speeding up the calculations.
\item Though SURF also creates octaves but it doesn't scales down the image, instead it changes the size of the box filter. (Scale Invariance)
\item Finding Keypoints: It uses Hessian Determinant for this purpose, which helps in expressing the local changes.
\item Then the Haar Wavelet responses are calculated again depending upon the scale similar to SIFT.
\item In the step above, each 4x4 sub region gives 4 values (Haar wavelet response), hence SURF calculates 64 dimensional feature vector.
\end{enumerate}

\subsection{Feature Matching}
The SIFT and SURF descriptors were matched using the FAST library for Approximate Nearest Neighbors (FLANN) \cite{muja2009fast}.

\subsection{Matching Accuracy and False Positives}

Matching accuracy is calculated by the following formula:

\begin{equation}
Accuracy = 1 - (\frac{False Positives}{Total Matches} *100)
\end{equation}
where false positives are the number of erroneously matched keypoints. 

False Positive is calculated by projecting the matched keypoints from reference image to the transformed image.

\section{Results\label{sec:results}}

The implementation was done using OpenCV 2.4.6 with Qt 5.0.2.  The OpenCV implementation of SIFT, SURF and FLANN are used for obtaining results. 
Additional Parameters for result generation:
\begin{enumerate}
\item Nearest neighbor distance: The minimum distance between descriptors was varied for matching as $t*min_distance$ where $t = \{2,5,10\}$.
\item	False Positives: Two neighbourhood sizes of $3x3$ and $9x9$ were used for marking a match as false positives.
\end{enumerate}
Each result considered in the following section has been obtained by averaging the results for individual images for corresponding matches.

\subsection{Scaling}

\begin{figure}[hbtp]
\centering
\includegraphics[width=\columnwidth]{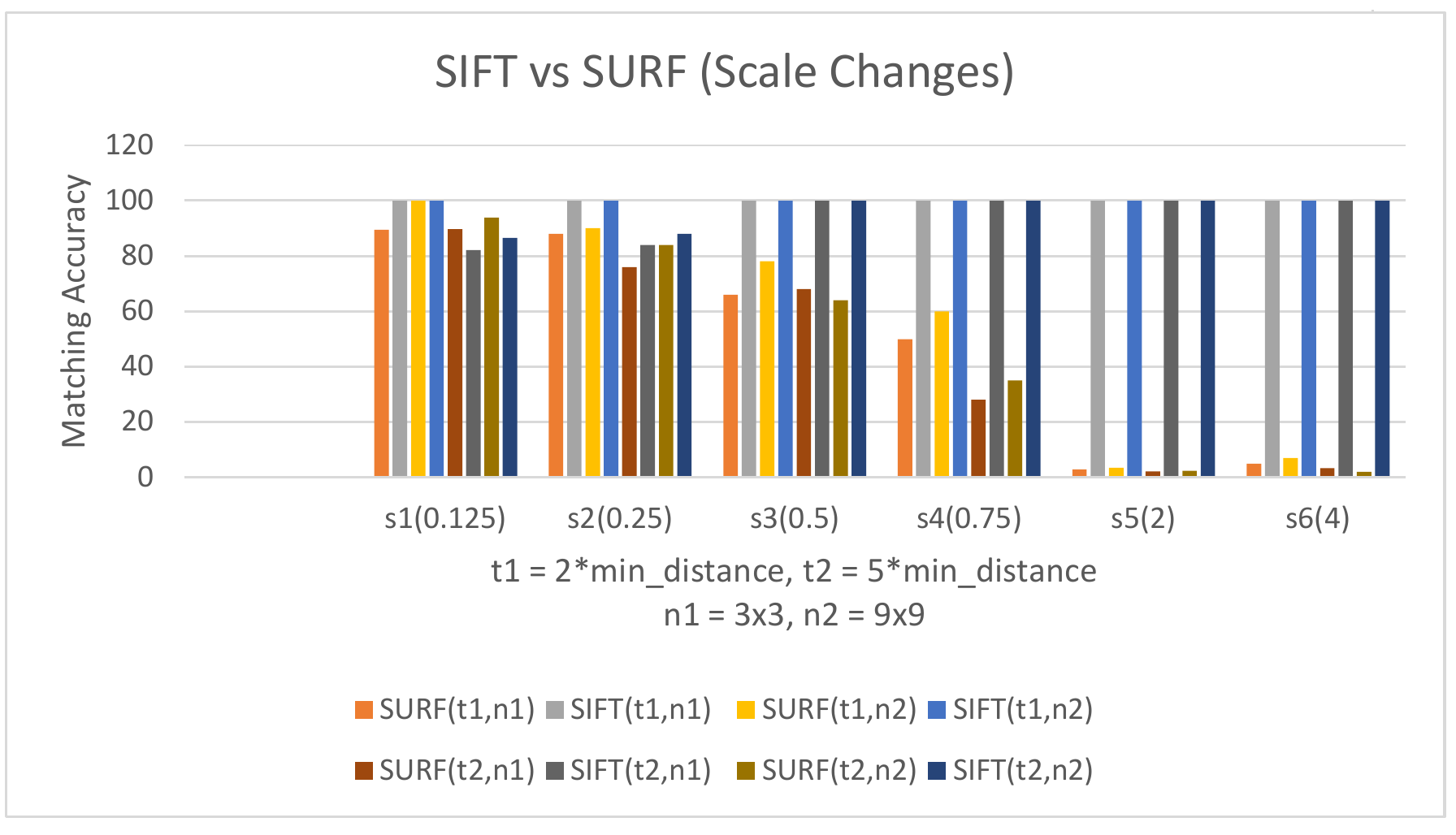}
\caption{Effect of scaling.}
\label{fig:6}
\end{figure}

The effect of scaling on the classification accuracy is shown in Fig \ref{fig:6}. As shown in the plot, as the minimum distance increases, the matching accuracy usually decreases. The reason for this is that when the threshold increases, more keypoints would be matched, but it also results in increase of false positives due to greater threshold. 

The plot also indicates that SIFT is more robust to scale changes than SURF indicated by the higher and consistent matching accuracy. It is also indicated from the plot of SURF (t1, n1) that SURF is more stable at lower scales than higher scales.

It was also expected that as the size of the neighborhood ($3x3$ to $9x9$) for finding false positives increases, the matching accuracy should have increased which is also evident from the plot.  For example, SURF $(t2, n2)$ has uniformly higher accuracy than SURF $(t2, n1)$. Hence, using two different neighborhoods wouldn’t be very significant in deciding the robustness of the techniques. Hence, rest of the results would consider only one neighborhood of $3x3$ for comparing results.

\subsection{Rotation}

\begin{figure}[hbtp]
\centering
\includegraphics[width=\columnwidth]{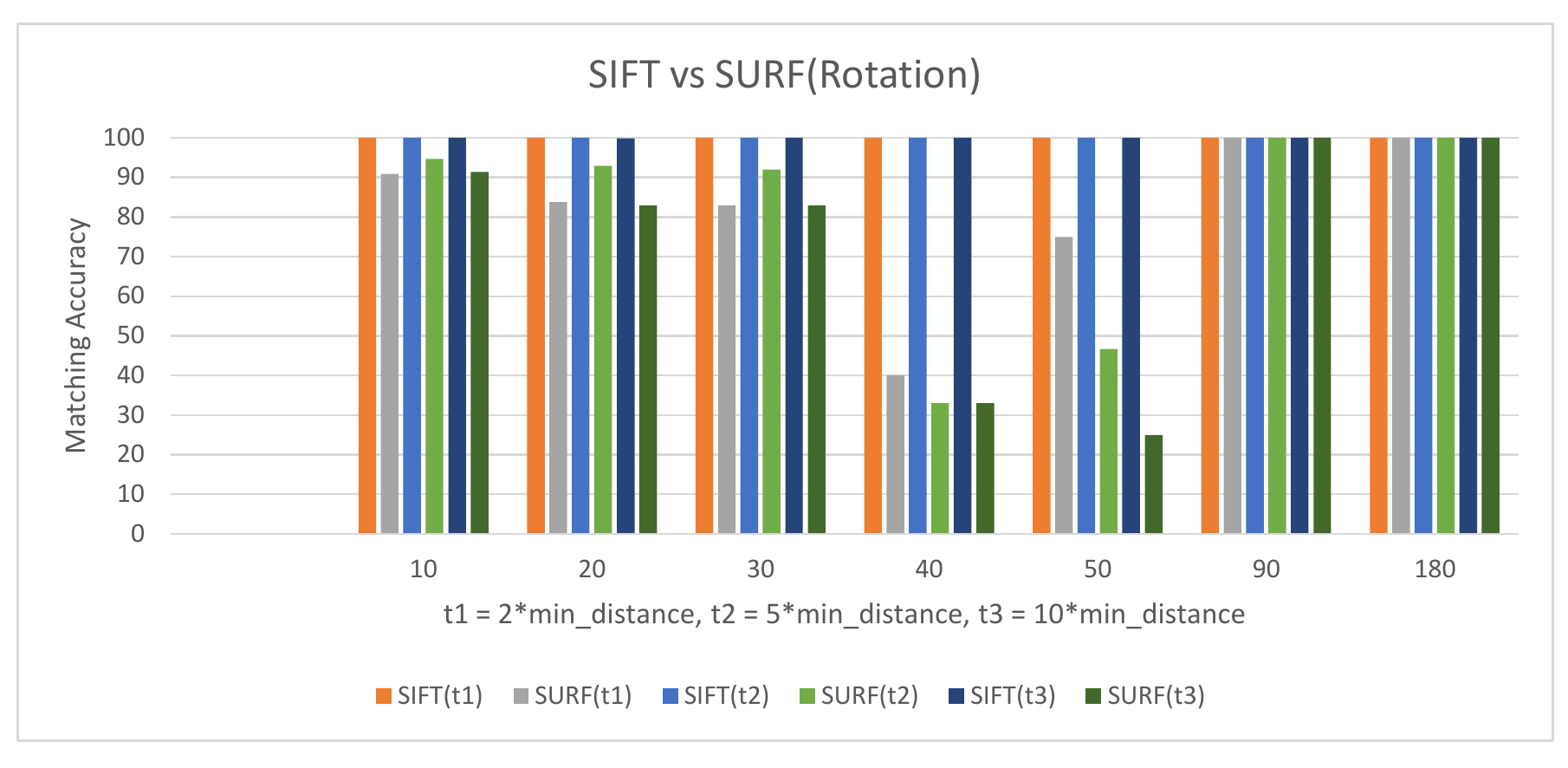}
\caption{Effect of rotation.}
\label{fig:7}
\end{figure}

The plot for rotation shown in Fig \ref{fig:7} compares the robustness at 10, 20, 30, 40, 50, 90 and 180 degrees respectively.

As indicated by the plot, SIFT outperforms SURF in consistency of the matches at various angles. It is shown that SURF’s rotation invariance decreases as the angle of deformation increases. But at 90 degrees and 180 degrees, the plot shows that SURF performs comparable matching efficiency to SIFT. This anomaly can be attributed to the type of images in the dataset. Since the images are of buildings consisting mostly of perpendicular and horizontal lines and the orientation of corners being symmetrical at doors, windows and edges of the building, the matching at 90 and 180 degrees finds the mostly the same keypoints which have stronger correspondence with the reference image than other orientations.

\subsection{Affine Transform}

\begin{figure}[hbtp]
\centering
\includegraphics[width=\columnwidth]{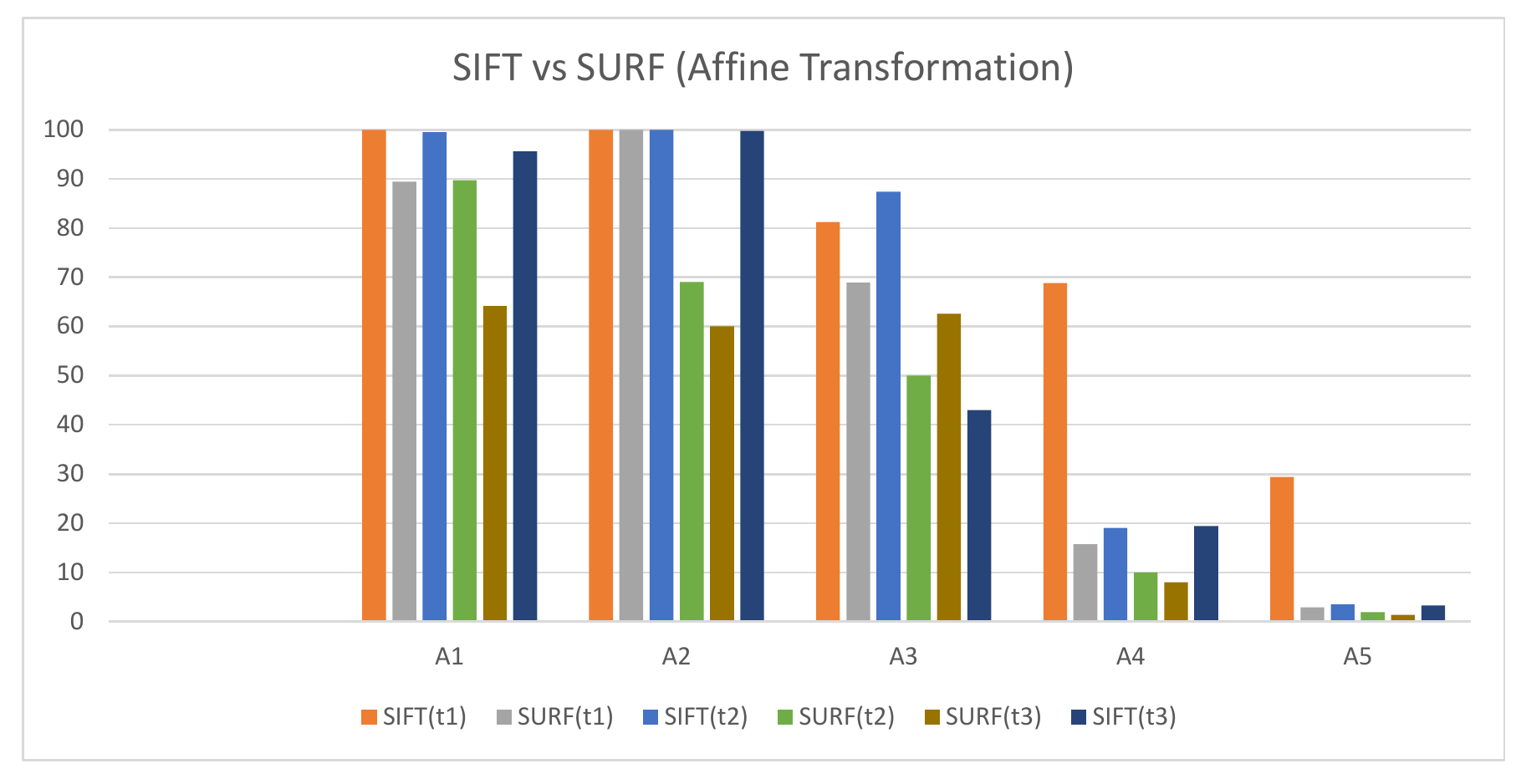}
\caption{Effect of Affine Transformations.}
\label{fig:8}
\end{figure}

The plot of affine transform (Fig \ref{fig:8}) shows the matching accuracy of SIFT and SURF with different affine transforms applied to the image and as shown in Fig \ref{fig:AffineTransforms}, where A1 corresponds to Fig \ref{fig:2b} and so on.

The plot indicates that SIFT outperforms SURF on invariance to Affine Transformation. For A1 to A3, SURF is pretty close to SIFT. This is also owed to the fact that A1 and A3 only have 10 to 30\% translation of the corner along both the axes, while A2 is essentially a scaled down version of reference image.  For A4 and A5, SIFT’s matching accuracy is much higher than SURF but is still not very accurate. Even by increasing the threshold (t1, t2, t3), the matching accuracy does not see any relative improvement as already showed and discussed in section for scaling.  A5 is the strongest affine transform, and the matching accuracy drops considerably when compared to other affine transforms. Hence, it can be said, that SIFT is invariant to only mild affine transformations i.e. which are essentially rotation or scale change across the axes.

\subsection{Perspective Transform}
\begin{figure}[hbtp]
\centering
\includegraphics[width=\columnwidth]{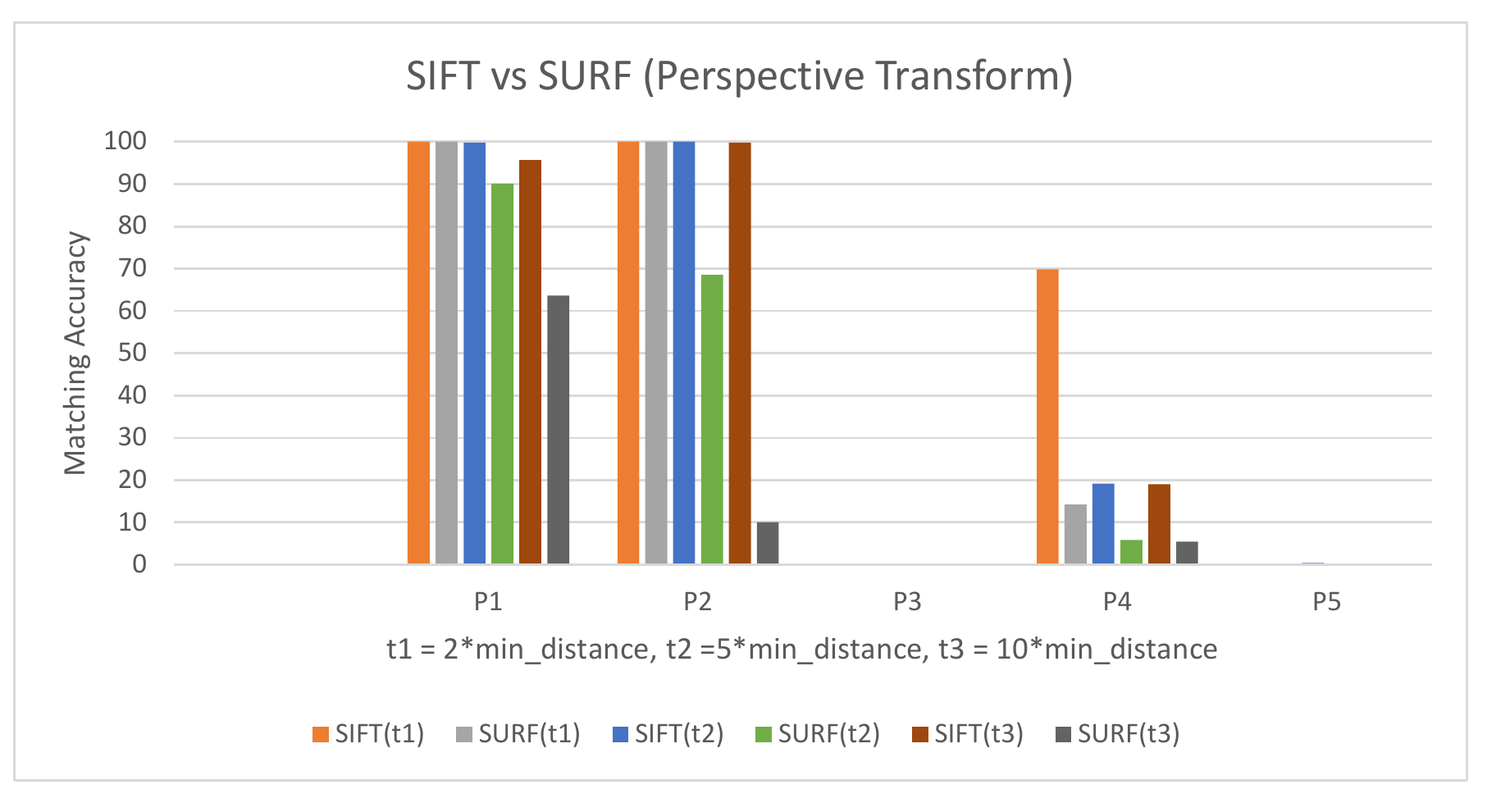}
\caption{Effect of Perspective Transformations.}
\label{fig:9}
\end{figure}

In Fig \ref{fig:9}, P1 to P5 correspond to Fig 3(b-f) respectively.   For P1 and P2, SIFT and SURF have comparable accuracies, owing to the fact that these are essentially translated and scaled down version of the reference image. 

For P4, SIFT has average matching accuracy, but as compared to P3 and P5, it has more restrictive transform. SIFT and SURF have very poor matching accuracy which can be explained from the fact that, these images correspond to more complex perspective transform as compared to others. Hence, it can be concluded that SIFT and SURF, both have extremely poor invariance to perspective transform when the viewpoint change is large while they have average matching accuracy in case of mild viewpoint change.

\subsection{Similarity Transform}
SIFT and SURF both have comparable matching capabilities for lower angles and scales while SIFT outperforming SURF for others. These results follow directly from discussion on Scale and rotation in-variances discussed above.

\begin{figure}[hbtp]
\centering
\includegraphics[width=\columnwidth]{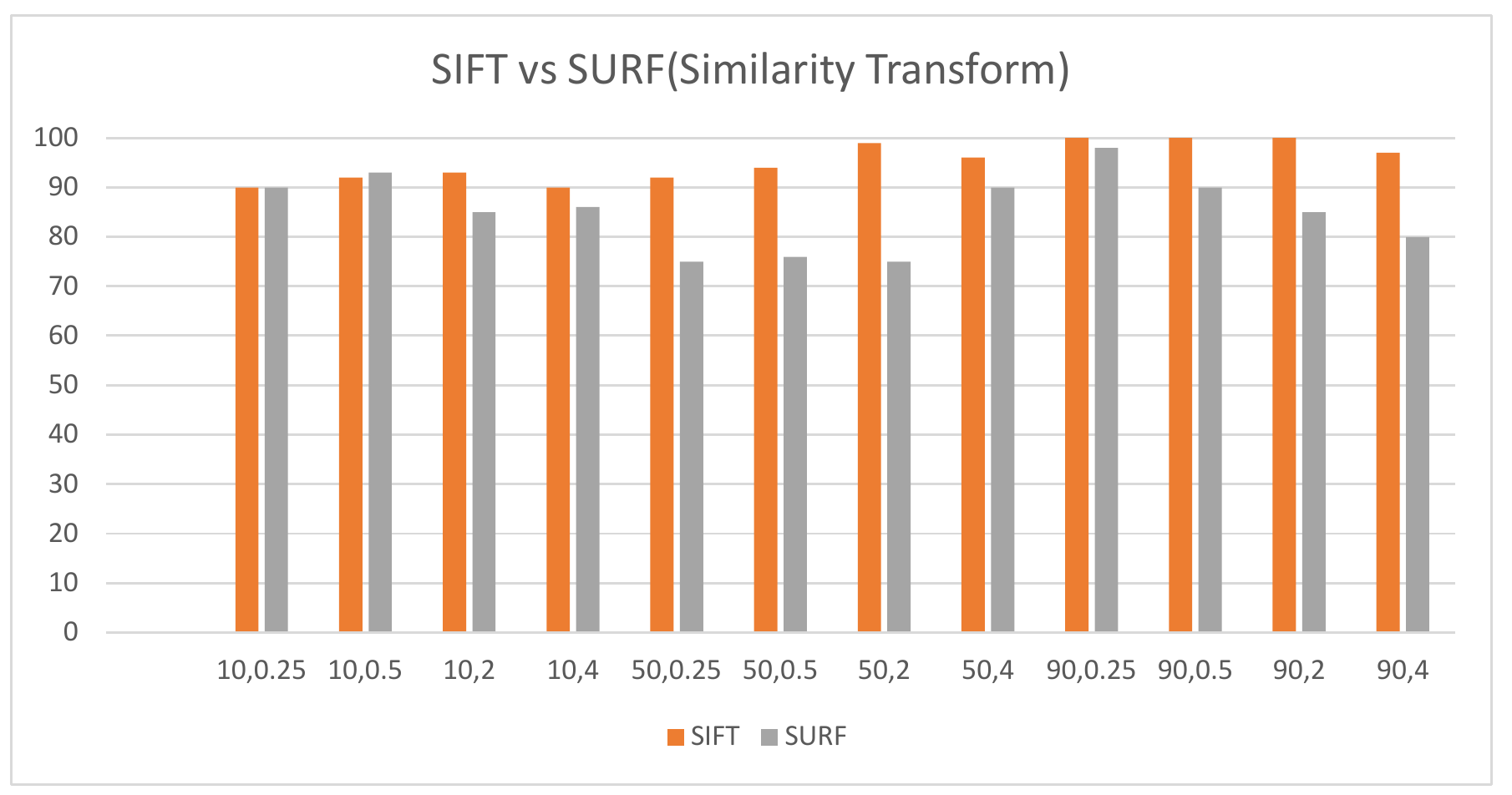}
\caption{Effect of Similarity Transformations.}
\label{fig:10}
\end{figure}

\section{Conclusion \label{sec:conclusion}}

Matching performance of SURF and SIFT were compared for trends in various transformations and anomalies arising in the results were analyzed.  It was found that SIFT outperforms SURF on almost all occasions while they both perform poorly for perspective transformation while partly being stable towards affine transformations. Existing studies on their comparison took only one parameter in consideration for concluding results. The performance of SIFT and SURF on specific scales and effect on them by increasing or decreasing scales was demonstrated, establishing that though SIFT and SURF both are invariant at lower scales, SIFT outperforms SURF on higher scales. Similar pattern was observed for rotation changes.

\section{ACKNOWLEDGEMENT}

I would like to thank Dr. Sumeet Agarwal, Assistant Professor, Indian Institute of Technology, Delhi and Dr. Hiranmay Ghosh, Principal Scientist, TCS Innovation Labs for guiding me through this work. 

\bibliographystyle{unsrt}
\bibliography{references}

\end{document}